\documentclass[10pt,twocolumn,letterpaper]{article}

\usepackage[pagenumbers]{cvpr} 

\usepackage{graphicx}
\usepackage{amsmath}
\usepackage{amssymb}
\usepackage{booktabs}
\usepackage{graphicx}
\usepackage[colorlinks=true,linkcolor=blue,urlcolor=blue,citecolor=blue,anchorcolor=red]{hyperref}
\usepackage{epstopdf}
\usepackage{pifont}

\newcommand{\cmark}{\ding{51}}%
\newcommand{\xmark}{\ding{55}}%
\usepackage{array, booktabs,ragged2e}
\usepackage[margin=1in]{geometry}
\newcolumntype{P}[1]{>{\centering\arraybackslash}p{#1}}
\newcolumntype{R}[1]{>{\RaggedLeft\arraybackslash}p{#1}}
\newcolumntype{L}[1]{>{\RaggedRight\arraybackslash}p{#1}}
\usepackage{xcolor}

\usepackage[capitalize]{cleveref}
\crefname{section}{Sec.}{Secs.}
\Crefname{section}{Section}{Sections}
\Crefname{table}{Table}{Tables}
\crefname{table}{Tab.}{Tabs.}

\def\cvprPaperID{26} 
\def\confName{MULA}
\def\confYear{2022}
\begin{document}

\title{AIMusicGuru: Music Assisted Human Pose Correction}

\author{Snehesh Shrestha, Cornelia Ferm{\"u}ller, Tianyu Huang, Pyone Thant Win, Adam Zukerman,\\ Chethan M. Parameshwara, Yiannis Aloimonos \\ \\
University of Maryland, College Park, MD, USA \\
{\tt\small \{snehesh,fermulcm,andy0412,pwin17,adamzuk,cmparam9,jyaloimo\}@umd.edu}
}
\maketitle

\begin{abstract}
Pose Estimation techniques rely on visual cues available through observations represented in the form of pixels. But the performance is bounded by the frame rate of the video and struggles from motion blur, occlusions, and temporal coherence. This issue is magnified when people are interacting with objects and instruments, for example playing the violin. Standard approaches for postprocessing use interpolation and smoothing functions to filter noise and fill gaps, but they cannot model highly non-linear motion. We present a method that leverages our understanding of the high degree of a causal relationship between the sound produced and the motion that produces them. We use the audio signature to refine and predict accurate human body pose motion models. We propose MAPnet (Music Assisted Pose network) for generating a fine grain motion model from sparse input pose sequences but continuous audio. To accelerate further research in this domain, we also open-source MAPdat, a new multi-modal dataset of 3D violin playing motion with music. We perform a comparison of different standard machine learning models and perform analysis on input modalities, sampling techniques, and audio and motion features. Experiments on MAPdat suggest multi-modal approaches like ours as a promising direction for tasks previously approached with visual methods only. Our results show both qualitatively and quantitatively how audio can be combined with visual observation to help improve any pose estimation methods.


\end{abstract}

\section{Introduction}
\label{sec:intro}

The future of education is going to be enhanced by AI \cite{selwyn2019should}. Online classes, self-practice, and journaling one's progress have become part of modern music education. Can AI enhance the experience by providing insights and analytics based on visual and aural observations for the students and the teachers \cite{chen2020ai}? Specifically, the art and skill of playing a musical instrument are learned through many iterations of lessons, practice, and feedback over multiple years. The teacher makes acute observations on how the student holds the instrument and moves to produce music. In addition, the teacher provides feedback on the movement corrections to produce the desired sound. The traditional approach involves motion capture systems, but it doesn't scale beyond research labs.

\begin{figure}
    \centering
    \includegraphics[width=0.8\columnwidth]{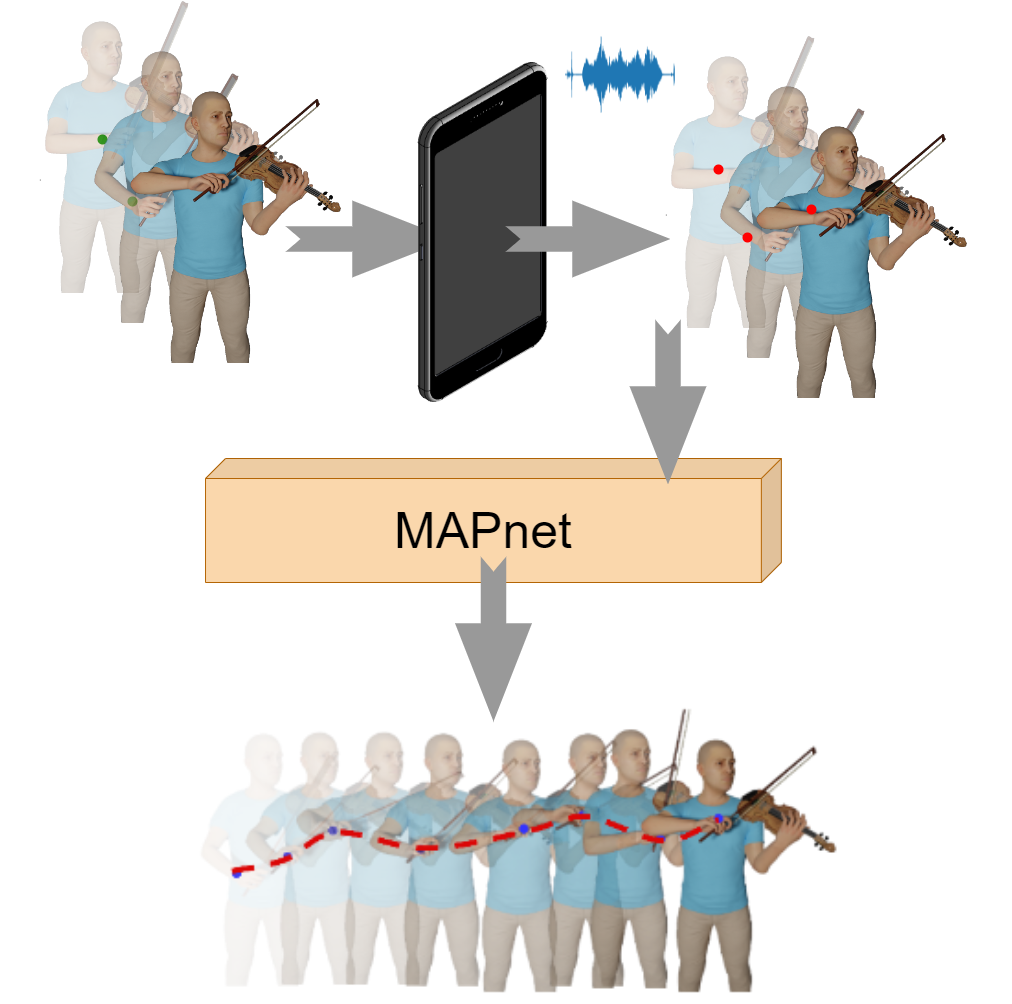}
    \caption{\textbf{AI Music Guru.} We present a new 3D violin playing music and motion dataset, MAPdat, which has rich motion capture ground truth, audio, and video of four master musicians. We also present MAPnet (Music Assisted Pose network), which generates a fine-grain motion model from sparse input pose estimate sequences and continuous audio. In this figure, a standard smartphone or a webcam can capture video at a low frame rate that MAPnet enhances to generate higher temporal resolution fine-grain pose estimates.}
    \label{fig:teaser}
\end{figure}

Modern visual pose estimation algorithms in 2D \cite{wei2016openpose_cpm, fang2017alphapose_rmpe} and 3D \cite{lugaresi2019mediapipe, pavllo:videopose3d:2019} have paved a promising path for using smartphones and webcams to estimate human pose. Approaches based on this technology are used already for assisting with large body movements such as in yoga, physical training, and golf \cite{difini2021human}. However, predicting poses for fine movements, such as in music training, is challenging where body movement is dictated by the audio generation. However, since the pose and audio signals are recorded at a different signal frequency (audio data are generated at a higher frequency than poses), current visual pose estimation models \cite{wei2016openpose_cpm, lugaresi2019mediapipe} are not equipped to handle the non-linear relationship between the music and the motion that produces them. In this work, we present the Music Assisted Pose network (MAPNet), which combines visual observations with temporally dense audio signals to produce accurate human pose as shown in Fig. \ref{fig:teaser}.

MAPnet learns the temporal distribution of pose and audio with the help of transformers. The features are then fused by a cross-modal fusion transformer that learns the distribution between pose and audio. Finally, the model is trained to predict poses at high temporal resolution in an auto-regressive sliding window fashion. Recent multi-modal pose and audio methods \cite{lee2019talking, karras2017audio,taylor2017deep,suwajanakorn2017synthesizing, shlizerman2018audio} aim at producing stylistically representative motion but not at predicting accurate pose. But well-recovered pose is essential for applications of music learning. To our knowledge, this is the first multi-modal approach that can predict correct human pose.

To train such a model, we need an appropriate dataset. While there are multi-modal datasets featuring people playing musical instruments \cite{li2018creating,sarasua2017musical,shlizerman2018audio}, they are collected in the wild without ground truth data. Collecting motion capture data of music players requires careful set-up and instrumented environments. In this work, we present a multi-modal audio-video dataset with precise ground truth human poses featuring people playing musical instruments. The authors of the TELMI \cite{Volpe2017-vr} database collected and released raw motion capture marker data along with its video. However, this data lacks kinematic human body joint modeling, body measurements, and calibration of the cameras for transformation from world 3D ground truth to image frame 3D coordinates. These make it difficult to repurpose them for machine learning tasks. We detail how we post-process and model the pose estimation challenges with the raw data to create this new dataset that we call Music Assisted Pose dataset (MAPdat), which could help this community towards a new research direction on multi-modal machine learning for music and motion.


To our knowledge, MAPdat is the first benchmark dataset for precise fine motor 3D pose estimation conditioned on music. In summary, our contributions are as follows:
\begin{itemize}
    \item We propose Music Assisted Pose network (MAPnet), which generates a high frame rate video and a precise pose estimation sequence from a low frame rate video and a low bandwidth music data.
    \item We introduce the Music Assisted Pose dataset (MAPdat) dataset containing music audio, high frame rate motion capture ground truth, and simulated pose estimation errors of advanced violin players.
\end{itemize}

\begin{table*}
    \centering
    \begin{tabular}{lcccccccc}

        \hline
\bf{Dataset}             & \bf{Type}    & \bf{Interaction}    & \bf{Fine}     & \bf{Pose}    & \bf{Estimates}        &  \bf{Audio}  & \bf{Video}   & \bf{Length (s)} \\
\hline
MoVi \cite{ghorbani2021movi}                & M & \xmark   & \xmark   & \cmark    & \xmark & \xmark        & \cmark        & 61,200 \\
HumanEva \cite{Sigal:IJCV:10b}          & M & \xmark   & \xmark   & \cmark    & \xmark & \xmark        & \cmark        & ~1,300 \\
Human3.6M \cite{6682899}          & M & \xmark & \xmark    & \cmark    & \cmark & \xmark        & \cmark        & 17,890 \\
\hline
TWH 16.2M \cite{lee2019talking}& S & \xmark & \xmark & $\sim$    & \xmark & \cmark        & \cmark        & 180,000 \\  
IEMOCAP     \cite{busso2008iemocap}       & S & \xmark & \xmark & $\sim$ & \cmark & \cmark & \cmark & 43,200 \\  
CreativeIT \cite{metallinou2016usc}              & S  & \xmark   & \xmark     & \cmark & \cmark & \cmark & \cmark & 28,800 \\
\hline
AIST++ \cite{li2021ai}              & D  & \xmark    & \xmark     & \cmark   & \cmark  & \cmark        & \cmark        & 18,694 \\
DwM \cite{tang2018dance}  & D  & \xmark    & \xmark     & \cmark    & \xmark & \cmark       & \xmark         & -- \\
GrooveNet \cite{alemi2017groovenet}           & D  & \xmark   & \xmark      & \cmark & \xmark & \cmark       & \xmark         &   -- \\
DanceNet  \cite{zhuang2020music2dance}          & D  & \xmark   & \xmark      & \cmark    & \xmark & \cmark       & \xmark         & --  \\
EA-MUD \cite{fan2011example}            & D  & \xmark    & \xmark     & \cmark    & \xmark & \cmark       & \xmark         & --  \\
PHENICX-conduct \cite{sarasua2017musical}    & C  & \xmark  & \xmark  & \cmark    & \xmark & \cmark       & $\sim$         & 3,420 \\
\hline
URMP    \cite{li2018creating}            & I  & \cmark & \xmark  & \xmark    & \xmark & \cmark       & \cmark         & 4,680 \\
QUARTET \cite{Papiotis:PhdThesis2016}        & I & \cmark & \cmark  & \xmark    & \xmark & \cmark       & $\sim$         & 1,742  \\
TELMI \cite{volpe2017multimodal}        & I & \cmark & \cmark  & \xmark    & \xmark & $\sim$       & $\sim$         & 8,625  \\
\hline
\bf{MAPdat (Ours)}       & \bf{I} & \cmark  & \cmark  & \cmark    & \cmark & \cmark       & \cmark         & \bf{120,690} \\
        \hline
    \end{tabular}
    \centering
    \caption{
    Dataset Comparison Table summarizes 3D human kinematic pose and audio datasets. Symbols (\cmark) means fully satisfies, (\xmark)  does not satisfy, and ($\sim$) partially satisfies the field topic. Dataset \textit{Type} include Motion (M), Speech (S), Dance (D), Conducting (C), and Play Instruments (I). \textit{Interaction} refers to people interacting with objects, thus making the dataset difficult for pose estimation due to occlusions, complex inter-body parts geometry, and joint deformation due to external forces. \textit{Fine} refers to the motion velocity and complexity, where slow-moving and large change is easier to observe. At the same time, fast-moving and small movements are much more challenging due to motion blur and pixel saturation. In the case of playing musical instruments, QUARTET and TELMI contain raw motion capture markers, audio, and video. URMP has audio and video. But they lack accurate 3D human joints, therefore lack 3D Pose Ground Truth \textit{Pose} and do not have GT to Video transformation calibration information making it difficult to compare with Pose \textit{Estimates} from video. MAPdat has full-body 3D human pose and simulated noisy pose estimates.}
    \label{table:dataset_comparison}
\end{table*}

\section{Related Work}
\label{sec:related}


\subsection{Pose Estimation with Audio}

The original pose estimation algorithms generate 2D skeletons from monocular image frames \cite{wei2016openpose_cpm, fang2017alphapose_rmpe}. However, due to limited expressiveness and ambiguity, 2D representations are often not sufficiently powerful for downstream tasks. Recently monocular 3D pose estimation approaches, often built on top of the 2D skeletons, have gained success \cite{lugaresi2019mediapipe, chen2020anatomy, gong2021poseaug, zhaoCVPR19semantic}. However, both 2D and 3D pose estimators suffer from issues such as jitter, joint inversions, joint swaps, and misses - issues extensively studied in \cite{lin2014microsoft, moon2019posefix}. In 3D, we observe additional challenges with scale, depth disparity, and joint drift. To overcome these issues, filter-based methods such as simple moving average, Kalman filters, and particle filters have been used \cite{Liu2018-ys} with some success, but they struggle to model highly non-linear motion trajectories.

Recently a new paradigm demonstrating the use of multi-modal data has emerged. It has been shown of tremendous success for generating poses based on audio data in dance sequences \cite{li2021ai, lee2019dancing, shlizerman2018audio}, for speech gesture generation \cite{li2021audio2gestures,ginosar2019learning}, and for learning new multi-modal features from visual and sound data that can be used with a variety of classic vision or audio tasks \cite{arandjelovic2017look}. 


\subsection{Pose and Audio Datasets}
While there are plenty of audio-video datasets for the areas of speech \cite{lee2019talking, busso2008iemocap, metallinou2016usc}, dance \cite{li2021ai, tang2018dance, alemi2017groovenet, zhuang2020music2dance, sarasua2017musical}, everyday actions \cite{ghorbani2021movi, 6682899, Sigal:IJCV:10b}, and music \cite{li2021ai, tang2018dance, alemi2017groovenet, Papiotis:PhdThesis2016, sarasua2017musical}, there are limited data when it comes to ground truth motion capture data on how people move while playing musical instruments. The table \ref{table:dataset_comparison} details the comparison of audio-video datasets.


There are other datasets with synchronized audio and motion capture data that are also used in gesture generation, such as AIST++\cite{li2021ai} and EA-MUD\cite{fan2011example}. These datasets have motion capture data from dances with synchronized music. However, in these scenarios, the audio is not directly produced due to human motion. Instead, people move in response to or with the anticipation of the beats of the music. As shown by prior work \cite{li2021ai}, they can be used for learning realistic human motions that are consistent with the motion styles. However, generation of such pose trajectories does not have a single unique solution, and approximations to the predictions are made. Therefore, the motion trajectories are not precise and do not conform to the real observations in the videos.

For the case of understanding the relationship between music production and the motion that produces them, we find four potential datasets \cite{jiangfcvid, li2018creating, Papiotis:PhdThesis2016, volpe2017multimodal}. FCVID \cite{jiangfcvid} contains YouTube videos from the wild hence lacking 3D pose ground truth. URMP \cite{li2018creating} was recorded in the lab but without a motion capture or multi-camera system, so it lacks the motion ground truth. QUARTET \cite{Papiotis:PhdThesis2016} consists of motion capture raw marker data; however, it only contains upper body parts, contains limited view angles from a single camera, and only contains 30 video sequences out of the 101 that were  recorded, making the data very limited. The TELMI \cite{volpe2017multimodal} dataset consists of 145 sessions with three different camera angles and full-body motion capture raw marker data of the body, violin, and bow. However, the motion capture marker needs kinematic human body model fitting to generate accurate 3D joint coordinates. TELMI does not have calibration data necessary for this mapping from ground truth to an image frame. Additionally, it has a limited number of subjects, making it difficult for downstream machine learning tasks. Thus we conducted extensive evaluation, synchronization, and post-processing steps to get the data to a machine learning-ready state. This is detailed in section \ref{sec:MAPdatdataset}.

\begin{figure*}
    \centering
    \includegraphics[width=0.9\linewidth]{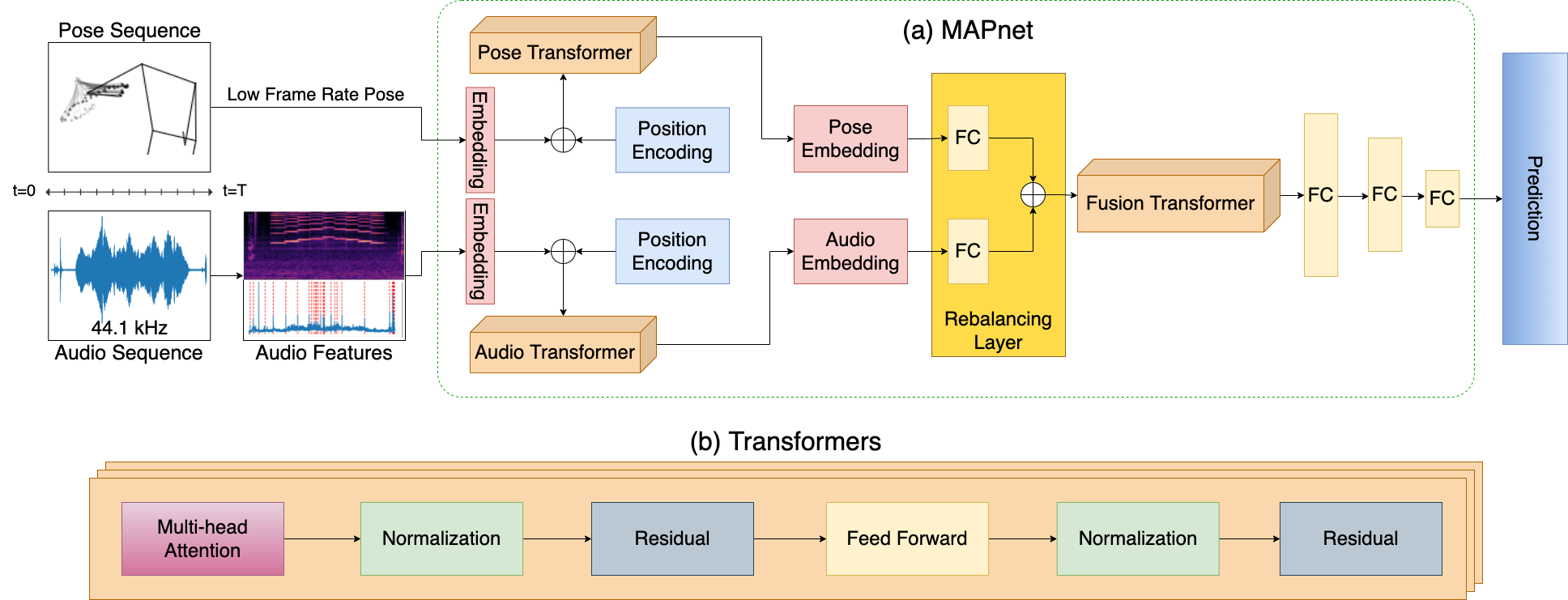}
    \caption{\textbf{Overview of the proposed MAPnet:} MAPNet consists of three transformers for pose, audio, and fusion respectively. The rebalancing layer is essential to encode pose and audio embedding for the fusion transformer to learn attention weights at different output fps to input fps ratios ($\mathbf{\tau}$). MAPnet takes 3 seconds sequence of Audio and 3D pose in varying low fps as input and will generate accurate 3D pose in high fps.}
    \label{fig:network_architecture}
\end{figure*}

\section{Method}
\label{sec:method}


\subsection{Overview}
As shown in Fig. \ref{fig:network_architecture}, we propose Music Assisted Pose Network (MAPnet) that takes temporally sparse and noisy pose skeleton data and refines them to generate temporally dense pose with the help of the audio. The primary building block of MAPnet are three transformers: (1) Pose Transformer and (2) Audio Transformer learn temporal attention from the input pose and audio signals. Since pose and audio embeddings are of different temporal dimensions, we use novel Rebalancing layers for both pose and audio embeddings. Further, the output pose and audio embeddings will be fused by the (3) Cross-Modal Fusion Transformer to learn the correspondences between pose and audio.


\begin{figure}
    \centering
    \includegraphics[width=0.8\columnwidth]{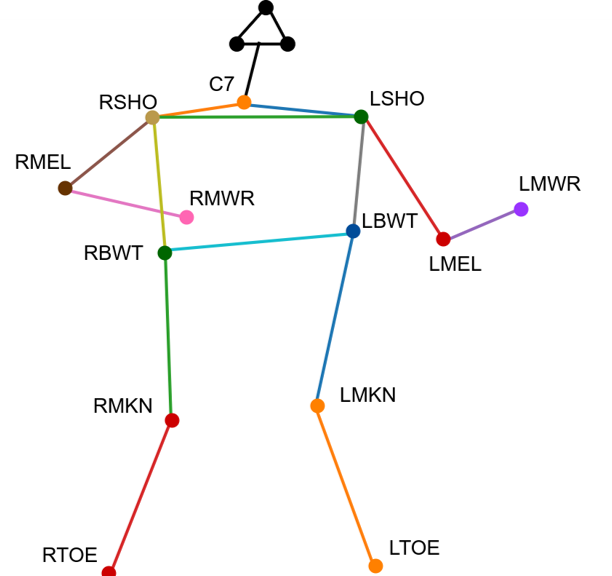}
    \caption{The above naming convention of  13 human body joints is based on TELMI. Each joint is represented in the Euclidean x, y, and z coordinate system. For example, C7 is the 7th cervical vertebra, RSHO and LSHO are right and left shoulders, RMEL and LMEL are right and left elbows, RMWR and LMWR are right and left wrists, RBWT and LBWT are left and right waist, RKNE, and LKNE are right and left knees, and RTOE and LTOE are the right and the left toes.}
    \label{fig:posefeatures}
\end{figure}

\begin{figure}
    \centering
    \includegraphics[width=0.7\columnwidth]{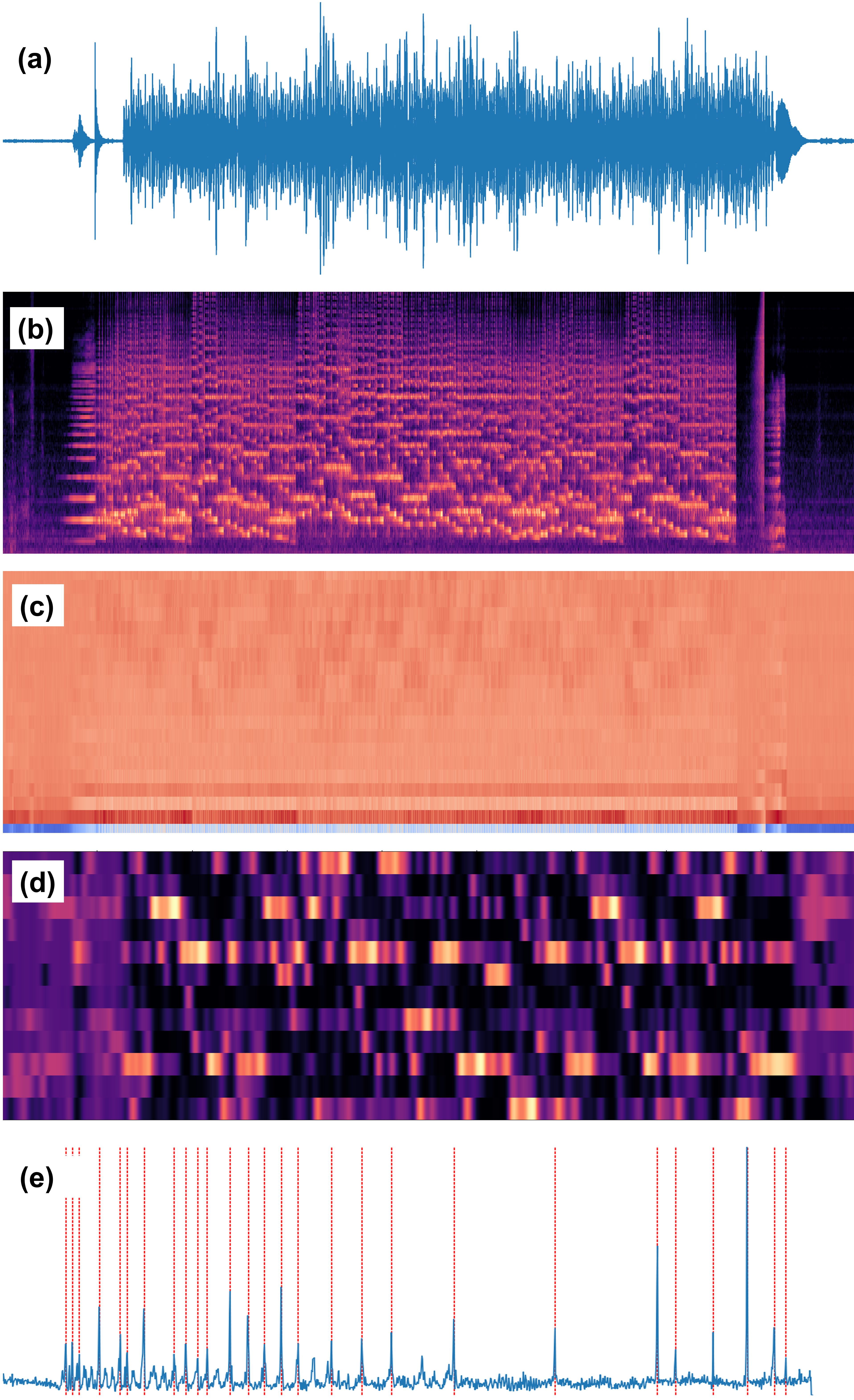}
    \caption{(a) Input audio waveform and (b) Mel-Power Spectrogram visualization. The Audio Input Features: (c) Mel-Frequency Cepstral Coefficients (MFCCs) (d) Chroma Energy Normalized (CENS) (e) Onset Strength and Peaks of Spectral flux onset strength envelope.}
    \label{fig:audiofeatures}
\end{figure}


\subsection{Preliminaries: Transformers}

Initially introduced in natural language processing for machine translation, the Transformer \cite{vaswani2017attention} is a very powerful technique employing multi-head attention (MHA), which allows a model to learn to pay attention to multiple salient features in sequential data. Transformers' fundamental building blocks are an MHA followed by a feed-forward (FF) layer with normalization and residual computed after each layer.

\subsection{MAPnet: Motion Assisted Pose Network}
As motion features, we directly take the sequence of normalized poses as shown in Fig. \ref{fig:posefeatures}. The input 3D skeletons have dimension $\mathbb{R}^{T_{in}\times13\times 3}$, where $T_{in}$ is the number of input frames and is equal to 3 seconds times the input frame rate. We have 13 joints for each person, and each joint is described by its  3D position (x,y,z). We flatten the last two dimensions to form our pose feature $P \in \mathbb{R}^{ T_{in}\times39}$.

\subsection{Audio Features}

Raw audio which are sampled at 44 kH are divided into 3 seconds sliding window with 1 second hop. Each 3 second audio is then divided into 150 time steps to generate 150 audio features. We compared raw audio and audio features used by other audio based networks \cite{li2021ai}. After our experiments, we carefully selected the following features: the 1-dim \textit{envelope} (Fig. \ref{fig:audiofeatures}e), 20-dim \textit{MFCC} (Fig. \ref{fig:audiofeatures}c), 12-dim \textit{chroma} (Fig. \ref{fig:audiofeatures}d), 1-dim \textit{one-hot peaks} (Fig. \ref{fig:audiofeatures}e), and 1-dim \textit{rms} to obtain a  35-dim music feature at 150 time instances $A \in \mathbb{R}^{ 150\times35}$.

Given the pose feature $P \in \mathbb{R}^{ B\times T_{in}\times39}$ and the audio feature $A \in \mathbb{R}^{ B\times150\times35}$, where B is the batch size, MAPnet first embeds the two features using fixed hidden size $H_1$ into $P \in \mathbb{R}^{ B\times T_{in} \times H_1}$ and $A \in \mathbb{R}^{ B\times 150\times H_1}$. 
The two embeddings are then fed into the Pose Transformer and the Audio Transformer, respectively, with positional encoding. The positional encoding ensures the temporal order of the concatenated features. \color{black}The output embeddings are then passed to the Rebalancing layer to reshape the embeddings into ${P \in \mathbb{R}^{ B\times H_2\times H_1}}$ and $A \in \mathbb{R}^{ B\times H_2\times H_1}$. These are then concatenated to get the combined  embedding $C \in \mathbb{R}^{ B\times 2H_2\times H_1}$ and sent to the fusion transformer without positional encoding.



We pass the output of the fusion transformer into three fully connected layers to get our output $ \in \mathbb{R}^{ B\times T_{out}\times 39}$, which gets reshaped into $\mathbb{R}^{B\times T_{out}\times 13\times 3}$ to calculate the loss with the ground truth, where $T_{out}$ is the number of output frames and is equal to 3 seconds times the output frame rate. The whole network with three transformers is learned in an end-to-end-manner.

\subsection{Reblancing Layer}
In our experiments, simply concatenating the modalities do not work, and the number of layers on individual transformers and fusion transformers make a huge difference. Therefore, our introduction of the rebalancing layer is vital to make the model work, for which we include ablation study (see table \ref{table:rebalance}). The rebalancing layer re-encodes the output from the individual pose and audio transformers. Doing this reduces the load for the fusion transformer to learn the attention mapping distribution of the pose and audio transformer embeddings.

\subsection{Loss Function}
We use the Mean Per Joint Position Error (MPJPE) as our loss function. 
MPJPE calculates the mean of the l2 loss between the ground truth joint position and the predicted joint position.
For our case, the MPJPE  is defined as:
$$
\mathcal{L}=\frac{1}{13} \Sigma_{n=1}^{13}\left\|j_{n}-\hat{j}_{n}\right\|_{2}
$$
where $j_{n}$ and $\hat{j}_{n}$ are the ground truth and estimated 3D joint coordinates of the n-th joint.



\section{MAPdat Dataset}
\label{sec:MAPdatdataset}

\subsection{Data Description}
The proposed MAPdat dataset is developed based on the publicly available database called TELMI Open Database \cite{volpe2017multimodal}. The TELMI dataset includes motion capture data, depth data, video data, and sound data of four master practitioners playing different violin techniques ranging from basic techniques such as controlling the bow weight while playing a violin to advanced techniques such as staccato articulation, sautille articulation, etc.

\begin{figure*}
    \centering
    \includegraphics[width=\textwidth]{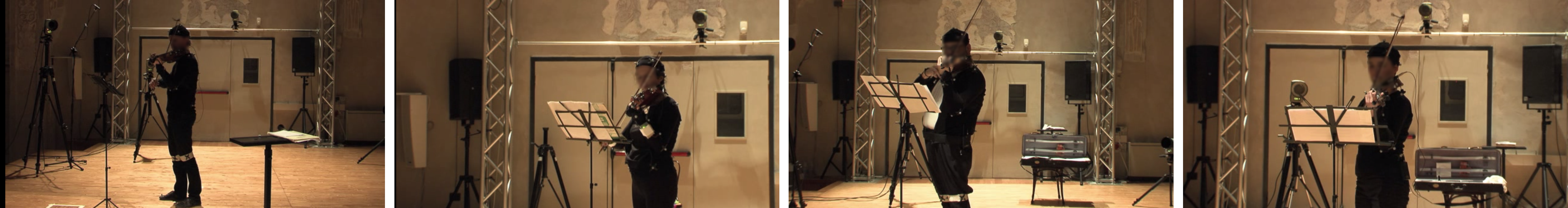}
    \caption{The four master practitioners are recorded in the TELMI dataset. These are included as part of the MAPnet video data.}
    \label{fig:telmeteachers}
\end{figure*}

The recordings are of master practitioners. Two to four of the master practitioners performed 41 different techniques as shown in figure \ref{fig:telmeteachers}. Although much significant research has been done by the original owners of the TELMI dataset, it was not in an ideal format to be used for machine learning.

Our current MAPdat is a prototype dataset and uses only a subset of TELMI. Since the  TELMI consortium owns the TELMI data, we are releasing our scripts to automate the download, post-process the original data, and generate machine learning-ready MAPdat data for the community to allow replication of our work.


\subsection{Dataset Preparation}
\label{sec:dataset_preparation}


While the TELMI dataset is rich, the dataset was built for a different type of research than our use case (e.g., \cite{Blanco2021-qw}). The video and motion capture data of the TELMI dataset is 50 frames per second, and the motion capture data is synchronized with audio and video data \cite{Volpe2017-vr}. The 32 motion capture markers of the body: ariel, right and left forehead, back-head, shoulder, back-shoulder, inner elbow, outer elbow, inner wrist, outer wrist, pinky, thumb, back waist, inner knee, outer knee, toe, and two points on the vertebrae (TS and T10) are included in the original data. We ignore markers that cannot be used for human body kinematic fitting with actual physical measurements such as the head and hand markers and do a kinematic fitting with useful markers using heuristic calculations of the clavicle, shoulders, elbows, wrists, hips, knees, and toes. From this, we compute a final 13 joints for a human body model as described in Fig. \ref{fig:posefeatures} consistent with the output of the video-based pose estimation approaches.


Although the motion capture data was synchronized with audio and video data, we found that the video, the audio, and the motion capture markers data had different lengths for many trials. Therefore, after they are temporally synchronized, we trim the data to have the same sizes. We validated the motion capture and audio-video synchronization by calculating the onset and offset of the music from the audio features and bow-violin distances and velocity from the motion capture data and doing a qualitative manual review. In TELMI data, data collection began and ended a few seconds before and after the master practitioners played violin. \color{black}Since those portions of data where no violin was being played are not relevant to our study, we have also cropped those parts before onset and after offset using the calculated bow-violin distances. We then transformed all the motion capture data with the left toe as the origin.

Each sample is normalized and then modeled with jitters as a gaussian noise and joint inversions as random swaps of joints to create ten variations, detailed in the next section. These synthetic noises are based on the most common types of challenges that pose estimators face \cite{Liu2018-ys}. These new 10x sets are randomly divided into 8:1:1 ratio for train:valid:test sets. The resulting sample and its corresponding audio samples are then re-sampled into three-second windows with a one-second hop as a sliding window. This results in 40,230 samples or 120,690 seconds of total data.



\subsection{Pose Error Characterization}
From our pilot studies, we find that the two most prominent forms of noise in 3D pose estimation are jitter and joint inversions, which was reinforced by \cite{lin2014microsoft, moon2019posefix}. Depending on the pose estimation network, jitter can vary in magnitudes, and joint inversion can occur with variation in the frequency of swaps and with which joints they get swapped. To emulate the noisy pose estimates of a monocular 3D pose estimator, we augment jitter and joint inversion to our ground truth motion capture data. We determined the noise parameters based on the average error from 3D pose estimates of MediaPipe \cite{lugaresi2019mediapipe} and VideoPose3D \cite{pavllo:videopose3d:2019} on the TELMI \cite{Volpe2017-vr} video data. The magnitude of Gaussian noise is based on the average distance of pose data from the ground truth motion capture data. While joint swaps are highly dependent on the camera viewpoint, we randomly distribute joint swaps based on the average occurrence from the total length of the video. In MAPdat, we modeled jitter as Gaussian noise distribution with a standard deviation of 300 mm. We modeled the distribution of joint swapping events over time as a Poisson random variable with an average of five swaps per minute. Once a joint swap occurs, the swap length is uniformly distributed between 0.5s and 3.0s. Using these parameters, we randomly generate ten variants of each video.

\section{Experiments}
\label{sec:experiments}

\subsection{Implementation Details}
We implemented all our methods in Tensorflow.
We used four NVIDIA RTX 1080 Ti GPUs to do our training and testing.
The  number of output frames $T_{out}$ across our experiments is 150 so that our model will predict 3D skeleton poses at 50 fps ($\tau$=1.0). We define $\tau$ as the ratio of output to input i.e. our output frame rate is 50, so the input frame rates at 50, 25, and 17 corresponds to $\tau$=1.0, $\tau$=0.5, and $\tau$=0.33. The input frame rate in the experiments was 50fps ($\tau$=1.0), 25fps ($\tau$=0.5), and 17fps ($\tau$=0.33), which correspond to same, one half, and one third of the output frame rate.
The  size $H_1$ and $H_2$ in the hidden layers used to embed the features was set to 160 and 150, respectively.
For all experiments we used in training a  batch size of 128 using the Adam optimizer \cite{kingma2014adam}.

To extract audio input features, we use audio processing toolbox Librosa \cite{mcfee2015librosa}. Most dance-based pose generation uses the beats feature as a cyclical marker for motions, such as foot-to-ground contact. In the case of the violin playing music, no distinct beats can be detected. Instead, we use changes in bow direction as the one-hot peaks and the RMS feature that better model the bow motion. The details of a sample waveform and audio feature have been visualized in Fig. \ref{fig:audiofeatures}.
\begin{figure}
    \centering
    \includegraphics[width=\columnwidth]{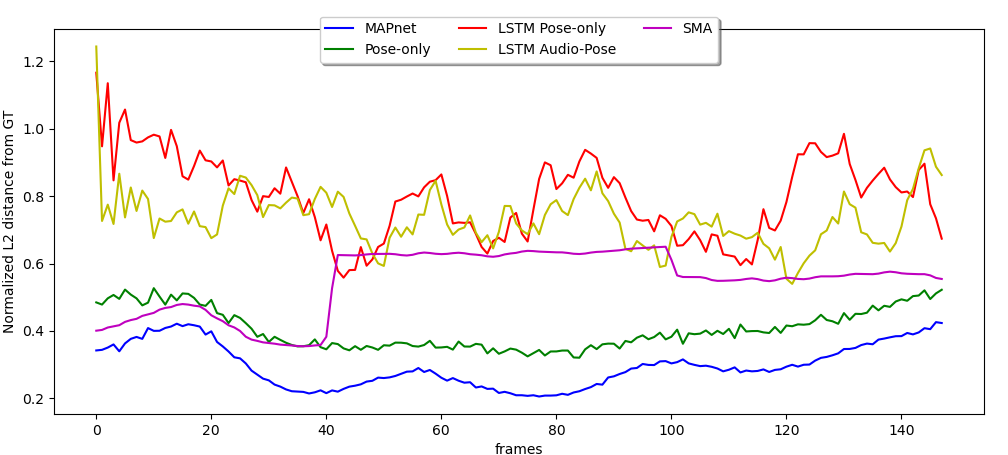}
    \caption{Error plot of normalized L2 distance between ground truth and prediction outputs at each frame. Blue: MAPnet predictions. Green: Pose-only transformer predictions. Magenta: Simple Moving average predictions. Red: LSTM pose-only predictions. Olive: LSTM Audio and pose predictions.}
    \label{fig:l2plot}
\end{figure}


\begin{figure*}
    \centering
    \includegraphics[width=0.8\textwidth]{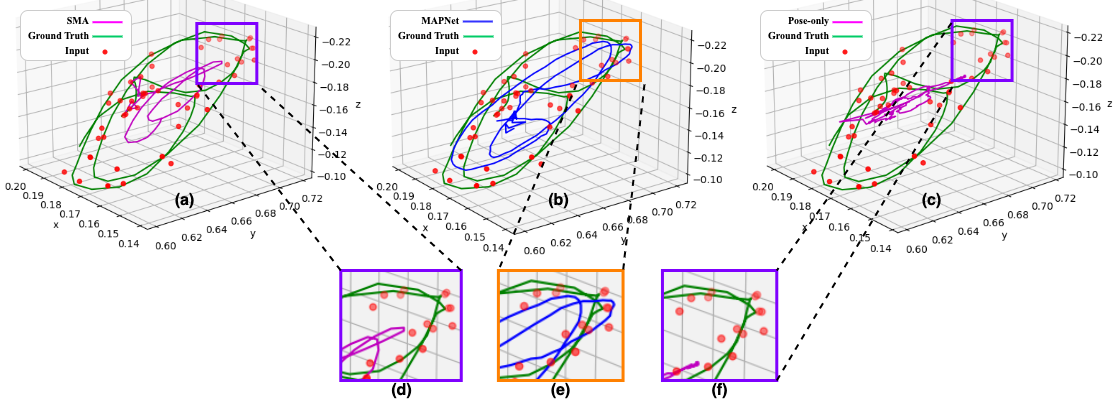}
    
    \caption{To demonstrate the difference in performance qualitatively, we plot the trajectory of the most salient joint, i.e., the right wrist, representing the bowing motion. In this plot, we compare the performance between (a) Simple Moving Average (SMA) (left), (c) Pose-only transformer (right), with (b) MAPnet model (center) in predicting the right wrist movement at $\tau$=0.33 input rate (17 fps). For clarity, we also show zoomed-in results of (d) SMA, (e) MAPnet model, and (f) Pose-only transformer.}
    \label{fig:simple_moving_average}
    \label{fig:poseVaudio}
\end{figure*}


\color{black}

\subsection{Analysis of Experimental Results}


    
    
    
    

We compare MAPnet with and without rebalancing with a Simple Moving Average calculation (SMA), a Long-Short Term Memory (LSTM) network using Pose only (LSTM Po), and one using Pose and Audio (LSTM PA), and to a Transformer based on Pose only (PoT). Table \ref{hand_stb} shows the quantitative results of our experiments. We used the Mean Per Joint Position Error (MPJPE) as our metric. For the Pose-only Transformer (PoT), we observed from the results that at 50 fps, the model struggles to filter out large jitter in the input data. PoT loses the motion trajectory data and deteriorates, as the number of input frame rate declines, at both $\tau$= 0.5 (25fps) input rate and $\tau$= 0.33 (17fps) input rate. In contrast, because MAPnet models the music features that have alignment with the actual motion trajectory, it filters the jitter and learns to model the joint swaps. Additionally, as we reduce the input frame rate (from 25 to 17), we do not see further deterioration but rather see a marginal improvement which can be explained by the transformer giving more attention to the audio signal to capture motion trajectories of very fine details, as can be observed in figure \ref{fig:poseVaudio}.

\begin{table}[t]
    \centering
    \resizebox{\columnwidth}{!}{
    \begin{tabular}{l|c c c}
        \hline
      \textbf{Method} &  $\mathbf{\tau=}$1.00  &  $\mathbf{\tau=}$0.50   &  $\mathbf{\tau=}$0.33  \\
       \hline
       SMA  & 292.31   & 303.63  &  311.20\\
       LSTM Po  & 345.86   & 341.80  &  338.17\\
       LSTM PA & 318.83   & 316.17  &  314.32\\
       PoT  & 35.49   & 43.16  &  41.65\\
       \hline
       MAPnet w/o Rebalance (Ours)  & \textbf{26.68} & 28.16  &  31.35\\
       MAPnet (Ours)  & 26.69 & \textbf{26.62}  &  \textbf{26.60}\\
        \hline
    \end{tabular}}
    \caption{Quantitative results of our experiments showing LSTM Pose-only (LSTM Po) and Pose and Audio (LSTM PA), Pose Only Transformer (PoT), and MAPnet correspond to the two networks we are comparing. The columns correspond to different input frame rates (given as the multiplication factor of the output frame rate). The numbers denote the MPJPE ($\downarrow$) in mm.}
    \label{hand_stb}
\end{table}


\begin{table}[t]
    \centering
    \begin{tabular}{l|c c c}
        \hline
      \textbf{Method} &  Fine & Gross & Inversions  \\
       \hline
       LSTM Po  & 385.17 & 310.62 & 273.72\\
       LSTM PA & 351.83 & 263.41  & 259.50\\
       PoT  & 46.27 & 44.43  &  29.01\\
       \hline
       MAPnet (Ours) & \textbf{27.76} & \textbf{22.29} & \textbf{19.05}\\
        \hline
    \end{tabular}
    \caption{Samples are categorized into three groups. Fine and Gross motion are based linearity of the data. Large motions and slow motions tend to be linear or piece-wise linear and therefore also the easiest for most models, including simple moving average to predict. Fine motions are highly non-linear, so we categorize and annotate these samples as the hard set. Inversions are when random joints get swapped with other joints. We categorize them as medium hardness as most non-linear (both Pose-only and MAPnet) perform well on this set. While simple filtering methods fail with very high MPJPE ($\downarrow$).}
    \label{motion_diff}
\end{table}


Simple Moving Average(SMA) is a traditional noise filtering method where the sum of a specified number of consecutive data is averaged to flatten out the noise. We have calculated the SMA on the input frames, and since it is a piece-wise linear noise-filtering system, we calculated the midpoints of the SMA results to generate higher frequency output. This method is not suitable for solving our problem of predicting motions at a high frame rate due to two reasons. Firstly, SMA is meant for linear systems, but the fine bow movements or movements of the right wrist are highly non-linear, as shown in Fig. \ref{fig:simple_moving_average} with the ground truth. Therefore, it cannot detect and learn the differences between fine bow movements and noise. Secondly, since it averages the input data, the prediction of the motions between two input frames is always predicted as linear, which is highly inaccurate for non-linear bow movements. As shown in Table \ref{hand_stb}, as the frame sparseness increases, the performance of a simple moving average algorithm deteriorates, while the MAPnet prediction, with the help of audio features, stays close to the trajectory of the ground truth. The results of the simple moving average from this ablation study were consistent with \cite{Liu2018-ys}.


Inspired from the LSTM network architecture presented in \cite{shlizerman2018audio}, we present the ablation study by replacing our transformer models with LSTM models. We observe that LSTM struggles to model the non-linearity between the pose and music. From table \ref{hand_stb}, the MPJPE error is exponentially high when compared to our MAPnet model, and therefore it fails to predict the accurate poses. We also observe the LSTM also struggles to learn different motion difficulty (see table \ref{motion_diff}). We anticipate this behavior is due to LSTM inherently depending on the sequential structure and being versatile in handling a single modality (either audio or pose). On the other hand, our transformer-base MAPnet is equipped to correlate multi-modal non-sequential data structures.

As shown in figure \ref{fig:poseVaudio}, at low input visual frame rate $\mathbf{\tau}=0.33$, during fine bow movements, The Pose-only model (Fig. \ref{fig:poseVaudio}a and \ref{fig:poseVaudio}d) and SMA (Fig. \ref{fig:poseVaudio}c and \ref{fig:poseVaudio}f) poorly predicts the motion of the wrist joints. However, the MAPnet model, due to coupling relations between the motion and the music, can generate more accurate predictions, as can be seen in the  Fig. \ref{fig:poseVaudio}b and \ref{fig:poseVaudio}e. We see that there is a significant error in the motion trajectory of the SMA and Pose-only model prediction, whereas there is a more substantial overlap between the MAPnet prediction and the ground truth.

\subsubsection{Ablation Studies}
\label{sec:ablation_studies}

\textbf{Rebalancing Layer: } We compared the effect of adding and removing the rebalancing layer, and we observed degradation of 6.31\% of per frame MPJPE metric and 17.64\% across frame MPJAE metric (see table \ref{table:rebalance}). This indicates that blindly concatenating the output of two transformers is not optimal.

\textbf{Early, balanced, or late fusion: } We also conducted early, balanced, and late fusion ablation (see table \ref{table:fusion}), which shows as much as 17.39\% difference in the MPJPE results. In this paper, our experiments suggest that pose only and traditional filtering methods do not work well, and our model outperforms them significantly.

\textbf{Results and Metrics: } In table \ref{table:fusion} and \ref{table:rebalance}, we include additional metric MPJAE (acceleration) to demonstrate temporal smoothness and correctness. The MPJAE is consistent with with MPJPE.


\begin{table}[t]
    \centering
    \begin{tabular}{l|c c}
        \hline
      \textbf{Ablation} &  MPJPE$(\downarrow)$ & MPJAE$(\downarrow)$ \\
       \hline
       Early-Fusion  & 32.20 & 44.37\\
       Balanced-Fusion    &28.22 & 41.53 \\
       Late-Fusion  & \textbf{26.60} & \textbf{37.51} \\
        \hline
    \end{tabular}
    \caption{We did ablation studies on when to fuse the pose and audio modalities. We conduct experiments in three settings, (1) Early-Fusion: 2 layers of Pose/Audio Transformer and 12 layers of Fusion Transformer (2) Balanced-Fusion: 7 layers of Pose/Audio Transformer and seven layers of Fusion Transformer (3) Late-Fusion: 12 layers of Pose/Audio Transformer and two layers of Fusion Transformer. Our result shows that the Late-Fusion strategy performs better in MPJPE and MPJAE.}
    \label{table:fusion}
\end{table}
\begin{table}[t]
    \centering
    \begin{tabular}{l|c c}
        \hline
      \textbf{Ablation} &  MPJPE$(\downarrow)$ & MPJAE$(\downarrow)$ \\
       \hline
       Without Re-balancing  & 28.39 & 45.55\\
       With Re-balancing  & \textbf{26.60} & \textbf{37.51} \\
        \hline
    \end{tabular}
    \caption{We also conducted experiments to evaluate the importance of the re-balancing layer in MAPnet. Removing the re-balancing layer will make a difference of 6.31\% MPJPE and 17.65\% MPJAE.}
    \label{table:rebalance}
\end{table}


\section{Limitations}
\label{sec:limitations}
The current approach and methods have limitations that suggest future work. The ground truth data has a limited number of joints, especially hand keypoints that were not recorded. The joints are not kinematically accurate and lack mapping between ground truth mocap joints and video due to the lack of calibration data that are necessary to compare with pose estimators results. The data has a limited number of subjects with wide variation in the number of samples from each subject. The data has advanced players, which does not include many sound variations in playing quality, instrument characteristics, and environmental effects. This data has only one instrument (violin) and may not generalize to other instruments. The current ground truth is limited to 50 fps limiting the upper bound of the fast motion. The ground truth motion complexity has been classified into gross motion, fine motion, and inversions. More analysis is needed into the nature of the issues.

The current approach relies on motions that are accompanied by sound —however, certain motions, such as when the bow hand moves away from the violin may not have sound to assist. The input is raw joints and five types of sound features. While this shows promise as one possible proof of concept, more exploration is needed for other motion and sound features. The current method relies on the network to figure out the mapping of each joint coordinates as a vector with sound features as vectors. More work is needed to provide a better coupling of motion and sound. The loss function does not consider the motion trajectory and, therefore, is limited in accurately modeling fine-fast-moving splines. Comparison to cross-datasets and cross-SOTA methods are needed to test for generalization.

\section{Conclusion}
\label{sec:conclusion}
This paper presents a method that leverages the audio features to refine and predict accurate human body pose motion models. We propose MAPnet (Music Assisted Pose network) for generating a fine-grained motion model from sparse input pose sequences and continuous audio. To accelerate further research in this domain, we also open-source MAPdata, a new multi-modal dataset of 3D violin playing. We hope this work will be useful to Computer Vision researchers, who can leverage the rich data from other modalities to pair it with vision in creative ways. Our results suggest that audio can be a rich resource to correct and fill in visual information between adjacent frames. Our work shows promising results, and it also creates opportunities for future researchers for cases like music, where there is a strong causal coupling between movement and audio.

{\small
\bibliographystyle{ieee_fullname}
\bibliography{root}
}

\end{document}